\documentclass{article}

\usepackage[preprint]{neurips_2021}

\pdfoutput=1
\usepackage[utf8]{inputenc} 
\usepackage[T1]{fontenc}    
\usepackage{hyperref}       
\usepackage{url}            
\usepackage{booktabs}       
\usepackage{amsfonts}       
\usepackage{nicefrac}       
\usepackage{microtype}      
\usepackage{xcolor}         
\usepackage{url}
\usepackage{amsmath}
\usepackage{amsthm}
\usepackage{graphicx}
\usepackage{cleveref}
\usepackage{caption}
\usepackage{subcaption}
\usepackage{placeins}
\usepackage{algorithm2e}

\title{Minimizing Communication while Maximizing Performance in Multi-Agent Reinforcement Learning}

\author{%
   Varun Kumar Vijay \\
   Intel Labs\\
   \texttt{varun.v.kumar@intel.com}\\
   \And
   Hassam Sheikh \\
   Intel Labs\\
   \texttt{hassam.sheikh@intel.com}\\
   \And
   Somdeb Majumdar\\
   Intel Labs\\
   \texttt{somdeb.majumdar@intel.com}\\
   \And
   Mariano Phielipp\\
   Intel Labs\\
   \texttt{mariano.j.phielipp@intel.com}
}

\begin{document}

\maketitle

\begin{abstract}
  Inter-agent communication can improve performance in multi-agent coordination tasks with shared goals. Prior works have shown that such communication protocols can be learnt using multi-agent reinforcement learning (MARL) and message-passing network architectures. However, these models adopt an unconstrained communication model, where an agent communicates with all other agents at every step, even when the task does not require it. In real-world applications, where communication may have constraints on bandwidth, power and network capacity, such a model may be too costly. In this work, we explore ECNet, a simple method of minimizing communication penalties while maximizing a task-specific objective in MARL. Our optimization pipeline adopts REINFORCE and the Gumbel-Softmax re-parameterization trick. We also introduce two techniques to stabilize training.
  First, we alternate training between episodes that do and do not use the communication penalty, preventing our models from turning off their outgoing messages. Second, we show that repeating previously received messages helps our models retain information and further improves performance. We show that we can reduce communication by 75\% without any loss of performance.
\end{abstract}
\section{Introduction}

Multi-agent reinforcement learning (MARL) has been applied to increasingly challenging domains such as Starcraft \cite{starcraft} and DOTA \cite{openai_five}. In Hide-and-Seek \cite{openai_hide}, MARL has been shown to produce complex emergent behavior, including the use of tools in order to accomplish goals. In many domains, it is possible for agents to communicate with each other over a network. By doing so, agents can discover performant joint strategies for achieving a shared goal.



In most real-world settings, communication must happen under constraints of bandwidth, power capacity, etc. In such settings, communicating agents have a dual objective - minimize communication and maximize a task objective. These opposing objectives create a challenging learning problem. A typical failure mode, as we observe and address in our experiments, results in agents shutting down communication at the start of learning before they have learnt to construct meaningful messages.

Prior works on learning efficient communications have significant limitations (more details in Section \ref{related_work}). 
In this work, we show that learnt communication models can be trained to minimize communication while still accomplishing a cooperative task with minimal or no performance loss. Our approach, Efficient Communication Net or ECNet, directly optimizes the weighted sum of the task reward and a communication cost by learning communication gates. We experiment with two methods of training these gates: the first is to use REINFORCE \cite{reinforce} in a manner similar to IC3Net \cite{ic3net}, the second is to employ the Gumbel-Softmax re-parameterization trick~\cite{gumbel} .

We find that it is necessary to employ two techniques to stabilize training. First, throughout training, we alternate between using the communication penalty and training without it to avoid failure modes like early termination of communication. Second, we find it helpful to forward a message from a previous time-step if no new message is received. Applying these techniques, we find that that both REINFORCE and Gumbel-Softmax-based gates are able to achieve efficient communication.

We also observe that communication resources can be further conserved by selecting only those recipients that can benefit from a message. We extend our gating system to generate a pairwise communication mask and show that the learnt mask captures the correct connectivity structure.

In summary, our key contributions in this paper are:
\begin{itemize}
    \item We show that we can minimize communication without loss of performance using a simple technique: jointly optimizing task reward and a communication penalty.
    \item We introduce two stabilization techniques that significantly improve performance.
    \item We show that we can also learn pairwise communication gates that capture the connectivity structure required to solve a task.
\end{itemize}

\section{Related Work}
\label{related_work}
\textbf{Multi-Agent Reinforcement Learning:} Our work is related to prior applications of reinforcement learning to solve multi-agent problems. We follow works such as \cite{emergent,openai_hide,openai_five,starcraft,ic3net} in parameterizing each agent's policy as a deep neural network and using a policy gradient algorithm to optimize the networks. The choice of training algorithm differs across the literature, from Proximal Policy Optimization \cite{ppo} to Multiagent Deep Deterministic Policy Gradients \cite{maddpg} and a combination of evolutionary optimization and policy gradients \cite{merl} . Similar to \cite{commnet,ic3net}, we use a fully synchronous version of the A3C algorithm \cite{a3c}, referred to as A2C. Our work is similar to~\cite{openai_five,ic3net,merl} in sharing parameters across the agents' networks. We focus on cooperative problems with partial observability, which requires a high degree of interaction between the agents.

\textbf{Multi-Agent Reinforcement Learning with Communication:} Prior work \cite{maddpg,commnet,tarmac} has shown that multi-agent reinforcement learning performance can be significantly improved by allowing a degree of communication between agents. The communication can be modeled either as part of a discrete action space~\cite{maddpg,rmaddpg}, or as a continuous hidden layer within the agents' networks~\cite{commnet,tarmac,sarnet}. We follow the latter approach. CommNet \cite{commnet}, TarMAC \cite{tarmac}, and SarNet \cite{sarnet} show that augmenting the communication mechanism with neural network modules such as attention and memory can help scale to more challenging tasks. We leverage these techniques by testing with both CommNet and TarMAC as the base architecture.

\textbf{Learning when to Communicate in Multi-Agent Reinforcement Learning:} A few recent studies have tackled the subject of deciding when to communicate in multi-agent settings. IC3Net \cite{ic3net} learns discrete communication gates using A2C in order to maximize task performance in mixed cooperative-competitive settings. We also experiment with A2C as a means of training gates, but our tasks are fully competitive and our training objective takes into account the cost of communicating. We note that the use of gating in IC3Net may be regarded as a partial solution to a flaw in their design which causes agents to leak information to competing agents; this point is made more fully in \cite{sarnet}.

Recent studies such as \cite{rmaddpg,avoidance} learn communication decisions with the objective of reducing the amount of communication. However, they both transmit predefined messages that are known to be useful, such as the observation vector, and therefore tackle a much simpler learning problem. Further, they cannot be applied to tasks in which the optimal messages are not known beforehand.

I2C \cite{i2c} is the work most similar to ours, learning both the messages and pairwise communication decisions. It differs from our work in several important ways. First, and most importantly, I2C does not use the delayed message delivery framework employed in previous models such as CommNet, IC3Net, and TarMAC. Instead, it uses a request-reply framework in which the entire round-trip takes place before the environment makes a transition. We note that this assumes fast communication latency, and may not always be practical in real-world settings. Second, I2C does not directly target reduced communication but obtains reduced communication as a by-product of optimizing for task performance. Third, the communication-gating network and the policies are trained in a two-phase manner rather than jointly.

Gated-ACML \cite{gacml} and ETC-Net \cite{etcnet} learn global gates in a two-phase manner. ETC-Net, which is the work most similar to ours, can be seen as ECNet-REINFORCE trained in a two-phase manner. However, this is a critical deficiency, as it does not allow the messages to be adapted to the final gating policy, limiting the reduction in communication. In fact, as we show in our experiments, it is possible to achieve the same communication reduction as ETC-Net using a random gating policy, requiring no learning. ECNet, by contrast, achieves a much lower level of communication at which learning is essential.

\textbf{Learning Latent Discrete Structures:} The problem of learning discrete latent units has been extensively explored in the literature on generative models. Two main algorithms are used for this purpose: REINFORCE \cite{reinforce} and the more recent Gumbel-Softmax reparameterization trick \cite{gumbel}, also named Concrete Distribution \cite{concrete}. We experiment with both of these methods. In the case of Gumbel-Softmax, we combine the reparameterized sampling with the straight-through estimator \cite{straight_through} to obtain samples that always lie on the one-hot simplex.

\section{ECNet: Efficient Communication Network}

\begin{figure*}[ht]
    \centering
    \includegraphics[width=0.99\linewidth]{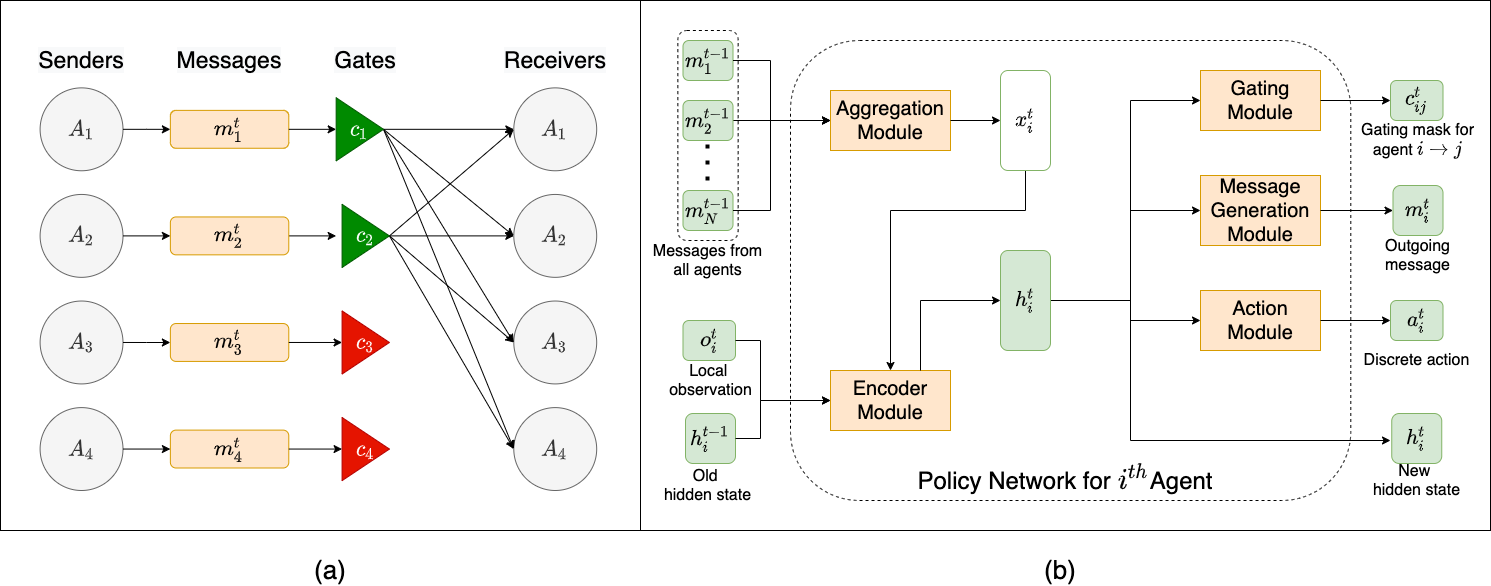}
    \caption{(a) The generated gates $c_i$ control communication between agents. (b) The operation of the policy network at the $i$th agent.}
    \label{fig:comm}
\end{figure*}

We first describe the policy network used in EC3Net focusing on the process by which agents generate actions at each time-step. We assume that the environment includes $N$ controllable agents. The policy network for agent $i$ receives three inputs at time-step $t$: the local observation $o^t_i$, its past hidden states $h^{t-1}_i$, and incoming messages from all agents $m^{t-1}_{j}$. It produces four outputs:
\begin{enumerate}
    \item $a^t_i \in |A|$, a discrete action
    \item $m^t_{i}$, the new message will be sent to all agents, including itself.
    \item  $c^t_{ij} \in {0, 1}$, a mask indicating whether agent $i$ will communicate with agent $j$
    \item $h^t_i$, new hidden states.
\end{enumerate}

The network consists of the following modules, as shown in \Cref{fig:comm}b:
\begin{enumerate}
    \item Message Aggregation Module: combines the incoming messages $m^{t-1}_{j}$ into an aggregated communication vector $x^t_i$.
    \item Encoder Module: combines the observation $o^t_i$, the aggregated communication vector $x^t_i$, and the previous hidden states $h^{t-1}_i$ and produces new hidden states.
    \item Message Gating Module: receives the current observation $o^t_i$ and the updated hidden states $h^{t_1}$ and computes a binary gate $c^t_{ij} \in {0, 1}$ controlling the communication from agent $i$ to agent $j$. As we discuss in subsequent sections, this module may be supplemented with additional inputs.
    \item Message Generation Module: runs after the Encoder Module and Message Gating Module and receives the updated hidden state $h^t_i$. It generates new outgoing messages $m^t_{i}$.
    \item Action Module: also receives the updated hidden state $h^t_i$ and generates discrete environment actions $a^t_i$.
\end{enumerate}

\subsection{Message Aggregation Module}
Since the architecture of the Message Aggregation Module is coupled to the Message Generation Module, we will describe it in the latter section.

\subsection{Encoder Module}
In all our experiments, the Encoder Module is an LSTM that receives its inputs as a concatenated vector.

\subsection{Message Gating Module}
We explore two forms of gating. The gates are generated at each time step.
\begin{enumerate}
    \item Global Gating. Each agent decides whether or not to communicate. The gate is generated by passing the current hidden state $h_{t}$ through a linear layer, followed by sampling from a Categorical or Gumbel-Softmax distribution. We show the operation of global gating in \cref{fig:comm} a.
    \item Pairwise Gating. Each agent decides whether or not to communicate with every other agent. The gate generator receives the current observation along with environment-provided information about the other agents. Here, the gating network is a two-layer MLP.
\end{enumerate}

For each type of gate, we explore two optimization methods:
\begin{enumerate}
    \item Gumbel-Softmax (GS). The reparameterization trick allows us to optimize the gate by backpropating through the network from the policy loss. In order to discourage unnecessary communication, we optimize a weighted sum of the policy gradient objective and a communication loss. For agent $i$, our objective is:
    \begin{equation*}
        J_i^{pg} - \lambda \sum_{i=1}^N \sum_{j=1}^N c_{ij}
    \end{equation*}

    The Gumbel-Softmax distribution computes a sample $y$ from log-odds $x\in \textbf{R}^n$ as:
    \begin{equation*}
    y=softmax((x + g) / \tau) \quad g \sim Gumbel(0, 1)
    \end{equation*}
    The sample $y$ can be reparameterized in terms of the input and a noise term and can therefore be optimized using backpropagation. $\tau$ is a temperature hyperparameter which controls the smoothness of the distribution. As the temperature goes to zero, the output of the distribution approaches a one-hot vector.

    We use a modified version of the Gumbel-Softmax called the Straight-Through Gumbel-Softmax, which produces samples that are one-hot while still being differentiable:
    \begin{equation*}
        y_{st} = stopgrad(onehot (argmax(y)) - y) + y
    \end{equation*}
    The straight-through estimator allows us to reduce the discrepancy between training, in which we use the Gumbel-Softmax, and testing, in which we take the maximum value.
    \item REINFORCE. Our second method of optimizing the gating mechanism is to treat the gates $c_{ij}$ as additional discrete actions and optimizing them using REINFORCE alongside the environment actions. The reward for the communication gates is a weighted sum of the environment reward and a communication penalty.
    \begin{equation*}
        R_{env} - \lambda \sum_{i=1}^N \sum_{j=1}^N c_{ij}
    \end{equation*}
    In order to account for the effect of communication on the recipient agents, we optimize the gates with a global reward, i.e. the mean of all agent rewards.
\end{enumerate}

We note that the combination of global gating and A2C is introduced in IC3Net. Our work adds the communication penalty as well as the stabilization techniques introduced in the next section.

\subsection{Message Generation Module}
We now describe the three communication architectures used in our experiments in terms of their generation and aggregation procedures. Two of the three communication architectures, CommNet and TarMAC, assume that they receive an incoming message from every other agent. If a message is absent, we fill in the message with a zero vector.

\subsubsection{CommNet}
The Message Generation Module in CommNet is a linear layer: $v_i = W h_i + b$. The values $v_j$ received from other agents are aggregated by summing and dividing by the total number of agents:
\begin{equation*}
    m_i = \frac{1}{N} \sum_{j=1}^{N} c_j v_j
\end{equation*}
In our experiments, the messages $v_j$ are vectors in $\mathbb{R}^{64}$.
\subsubsection{TarMAC}
TarMAC divides the message into two components, a key $k_i$ and a value $v_i$. In our experiments, we use$k_i \in \mathbb{R}^{16}$ and $v_i \in \mathbb{R}^{32}$. Each is generated using a linear layer: $k_i = W_k h_i + b_k$, $v_i = W_v h_i + b_v$.
TarMAC's Message Aggregation Module uses attention to focus on messages that are important to a receiving agent. Each agent locally generates a query vector using a linear layer: $q_i = W_q h_i + b_q$.
The query is multiplied against the incoming keys and passed through a softmax function to produce normalized attention weights:
\begin{equation*}
    \alpha_{ji} = \frac{\exp{c_{ji} (q_i^T k_j})}{\sum_{j'=1}^{N} \exp{c_{j'i} (q_i^T k_{j'})}}
\end{equation*}
The aggregated vector is a sum of the incoming values multiplied against the attention weights.
\begin{equation*}
    m_i = \sum_{j=1}^{N} c_{ji} \alpha_{ji} v_j
\end{equation*}
\subsubsection{TarMAC-Sigmoid}
In some of our experiments, we found that TarMAC had poor performance or failed to train. Empirically, we found improved performance we a modified architecture in which each message was processed independently, while retaining the ability to match queries against incoming keys while processing each message independently. Further, the modified architecture does not require knowledge of the total population size. TarMAC-Sigmoid differs from TarMAC only in its computation of attention weights. If a message is received by agent $i$ from agent $j$, the attention weight is $ \alpha_{ji} = \sigma(w_{scale} \times (q_i^T k_j) + b_{scale}$.
If no message is received, the attention weight is set to zero, i.e. the message is not added to the aggregated vector. $w_{scale}$ and $b_{scale}$ are trainable scalars. $\sigma$ is the sigmoid nonlinearity.

\subsection{Action Module}
The action module takes as input the current hidden state $h^t_i$. The action is computed using linear layer followed by sampling from a categorical distribution.

\section{Stabilization Techniques}
We introduce two stabilization techniques that prevent model performance from collapsing when training with both task reward and a communication penalty.
\begin{enumerate}
    \item \textbf{Multitask Training:} Our first stabilization method is to use construct two training tasks from environment: the first uses the communication penalty while the second has no communication penalty. We divide training episodes evenly between the two. A binary flag indicating whether the current episode uses the communication penalty is provided to the agent as part of its observation. This allows the agent to keep its communication gates open on the episodes that do not use the penalty, and therefore learn meaningful messages.
    \item \textbf{Message Forwarding:} We also hypothesize that performance may be limited by the inability of agents to retain information from previously received messages. In order to alleviate this problem, we experiment with repeating the most recently received message.
\end{enumerate}

\section{Experiments}

\begin{figure*}[ht]
    \centering
    \includegraphics[width=0.95\linewidth]{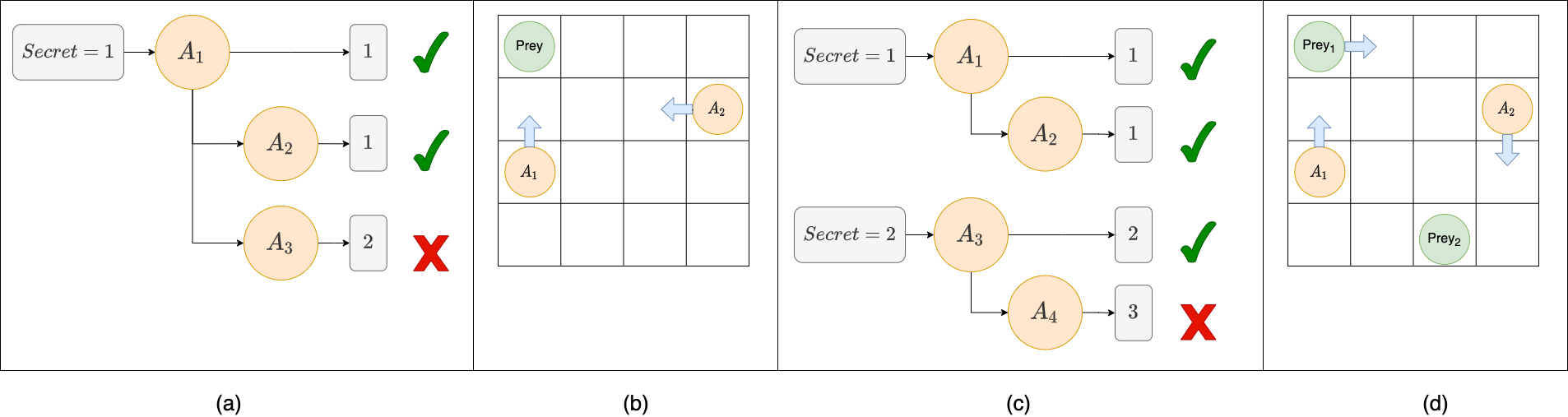}
    \caption{Illustrations of our three environments. (a) Secret: $A_1$ is given a secret and must broadcast it to the other agents. (b) The agents search through a partially available grid to locate the prey. (c) Secret Pairs: $A_1$ and $A_3$ need to send secrets to $A_2$ and $A_4$ respectively.}
    \label{fig:envs}
\end{figure*}

We test our models on four tasks in which communication is essential for obtaining good performance. The goal of our experiments is to assess how effective our methods are at decreasing communication overhead, while minimizing performance loss. In each of our experiments, we performed hyperparameter tuning to find communication penalties that achieve the desired performance and communication levels. 

\subsection{Secret}
The environment consists of 5 agents acting for 20 steps. The agents choose from 20 actions at each step. One of these actions is the secret and yields a reward of 1, while the others yield reward 0. The secret is fixed for the duration of the episode. It is given to one of the agents as its observation; the other agents need to communicate with the agent possessing the secret in order to solve the task.

In order to obtain maximum reward for all agents, the secret-holder needs to send only one message, while the other agents do not need to communicate. Our experiments investigate whether it is possible for agents to approach this protocol through reinforcement learning.

\subsection{Predator-Prey}
Our second environment tests the ability of our models to tackle a larger-scale task. This environment consists of 10 agents that must reach a single, stationary prey, located in a 20x20 grid, within 80 steps. The prey yields a reward of 1 per step to each occupying agent. The agents only have visibility of the four adjacent grid cells to their position. Communication is therefore essential in order to locate the prey quickly. Unlike in Secret, it is difficult to specify a priori when agents should communicate. If an agent is the first to reach the prey, it should probably inform its teammates; however, agents might also want to communicate information about regions in which the prey is absent.

We measure performance using a success metric: an episode is successful if all agents are simultaneously occupying the prey's grid cell at the same time. We note that this is a considerably more demanding metric than the episode return, which can be significant even when some agents do not reach the prey.

\subsection{Secret Pairs}
Our third environment tests the ability of agents to learn pairwise communication decisions. It involves 6 agents acting for 20 steps, choosing from 5 actions at each step. As in Secret, the agents must choose the correct action in order to obtain a reward of 1. Here, the agents are divided into pairs, identified by a pair id, and one agent in each pair is given the secret. Further, the secret changes every 5 timesteps.

The solution to this environment demands multiple messages to be sent, one for each change in the secret. However, the pattern of connectivity required is sparse: only the secret-holder in each pair needs to communicate, and it only needs to send a message to the other agent with the same id.

\subsection{Dynamic Cooperative Navigation}
Our final environment is the dynamic cooperative navigation environment described in \cite{etcnet}. This environment consists of two agents that each need to navigate to a destination. However, the destination is only provided to the other agent, making communication essential. Further, with a low probability, the destinations can move at each timestep, testing the ability of agents to respond dynamically.
\section{Results}
\subsection{Secret}
\begin{table*}[ht]
    \small
    \centering
    \begin{tabular}{lrrrrr}
    \toprule
    & \multicolumn{1}{c}{\textbf{Penalty}} & \multicolumn{1}{c}{\textbf{Sender Return}} & \multicolumn{1}{c}{\textbf{Receiver Return}} & \multicolumn{1}{c}{\textbf{Sender Comms}} & \multicolumn{1}{c}{\textbf{Receiver Comms}} \\ \midrule
    {TarMAC-Sigmoid} & {0.0} & {19.72$\pm$0.42} & {17.93$\pm$0.71} & {76.00$\pm$0.00} & {76.00$\pm$0.00} \\[0.05in]
    {IC3Net} & {0.0} & {19.24$\pm$0.50} & {18.08$\pm$0.31} & {63.88$\pm$3.93} & {46.11$\pm$8.93} \\[0.05in]
    {ECNet-REINFORCE} & {0.1} & {19.82$\pm$0.27} & {17.12$\pm$1.58} & {6.47$\pm$5.50} & {0.00$\pm$0.00} \\[0.05in]
    {ECNet-GS} & {0.01} & {19.98$\pm$0.03} & {12.68$\pm$2.57} & {3.38$\pm$1.54} & {0.07$\pm$0.10} \\[0.05in]
    \bottomrule
    \end{tabular}
    \caption{Task Return and Number of Messages in Secret with TarMAC-Sigmoid. We compare our method, ECNet, to baselines that do not use gating and IC3Net, which uses gating but does not use a communication penalty. ECNet-Reinforce performs on par with prior methods while using fewer than 7 sender messages and no receiver messages. In comparison, prior methods send over 50 messages. Results with all architectures are available in the Appendix.}
    \label{tab:secret_results_brief}
\end{table*}

\begin{figure*}[ht]
    \centering
    \includegraphics[width=0.9\linewidth]{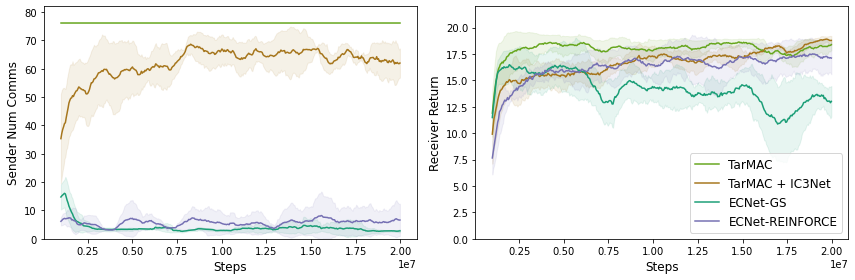}
    \caption{Performance and Communication in Secret using TarMAC-Sigmoid}
    \label{fig:secret_results}
\end{figure*}
In \Cref{fig:secret_results}, we show the performance and communication levels of various gating methods with the TarMAC-Sigmoid architecture, which generally performs the best. We see that the model that does not gate communications obtains near-perfect task performance, but uses a large number of messages. In contrast, the models that use global gating have somewhat lower task performance but use vastly fewer messages. The best performing gated model is ECNet-REINFORCE, which performs on par with the non-gated models but communicates only 10\% of the time. We observe that ECNet-GS tends to perform worse than ECNet-REINFORCE. In addition, we note that TarMAC is the worst performing architecture, while our modified TarMAC-Sigmoid model is the best, lending support to our hypothesis that the softmax operation over attention scores might interfere with gating.

In order to ascertain whether the task reward might be sufficient to reduce communication, we implement the setup described in IC3Net, optimizing the global gate with reinforce without using a communication penalty. We see that while these models do not always communicate, they still send a large number of unnecessary messages. Therefore, the communication penalty is essential.

\begin{figure*}[ht]
    \centering
    \includegraphics[width=0.90\linewidth]{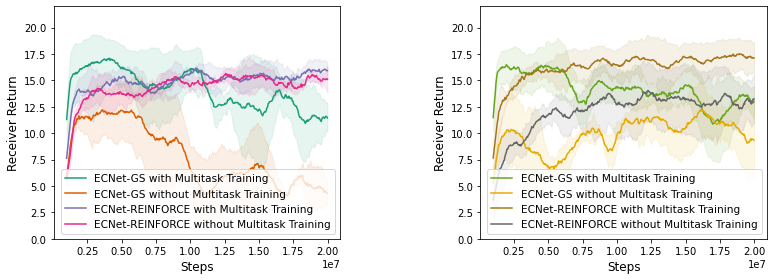}
    \caption{Effect of Multitask Training in Secret with CommNet (left) and TarMAC-Sigmoid (right)}
    \label{fig:secret_50}
\end{figure*}

We also use the Secret environment to investigate the two stabilization techniques introduced above: multitask training and message forwarding.~\Cref{fig:secret_50} shows the effect of multitask training on the receiver return. We observe that adding multitask training consistently improves performance over the baseline across two architectures and optimization methods. In the case of CommNet with ECNet-GS, it is crucial in preventing the communication mechanism from switching off.

\begin{figure*}[ht]
    \centering
    \includegraphics[width=0.9\linewidth]{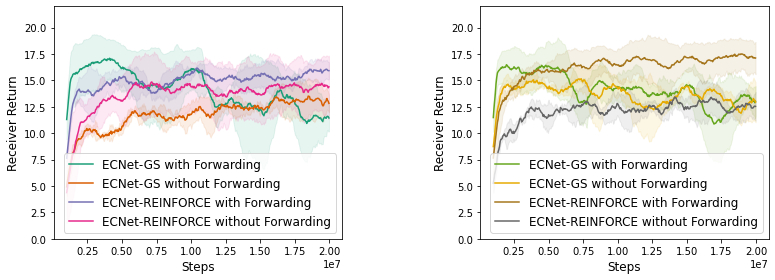}
    \caption{Effect of Message Forwarding in Secret with CommNet (left) and TarMAC-Sigmoid (right)}
    \label{fig:secret_forwarding}
\end{figure*}

Message forwarding (\Cref{fig:secret_forwarding}) has a less dramatic impact but still provides a performance gain in most cases. In particular, the best-performing model is Tarmac-Sigmoid with message forwarding. Therefore, in subsequent experiments, we use both multitask training and message forwarding.

\subsection{Predator-Prey}

In Predator-Prey, we similarly observe that our methods are effective at achieving a more efficient inter-agent messaging protocol. Communication is essential for this task: an independent model with no inter-agent communication is only successful in 14\% of episodes, while the best-performing model, TarMAC, is successful in over 80\%. For both ECNet-GS and ECNet-REINFORCE, we experiment with two penalty settings: a low-penalty setting, which reduces communication while retaining performance, and a high-penalty setting, which limits communication more severely but loses some performance. The best performing low-penalty model, TarMAC with ECNet-GS and penalty 1e-3, has over 80\% success but sends messages on only 25\% of steps. In contrast, the ECNet-REINFORCE models with high-penalty have 30\% success but communicate only a few times each episode. These results indicate how the communicate penalty can be used to control the trade-off between task performance and communication.

\begin{figure*}[ht]
    \centering
    \includegraphics[width=0.9\linewidth]{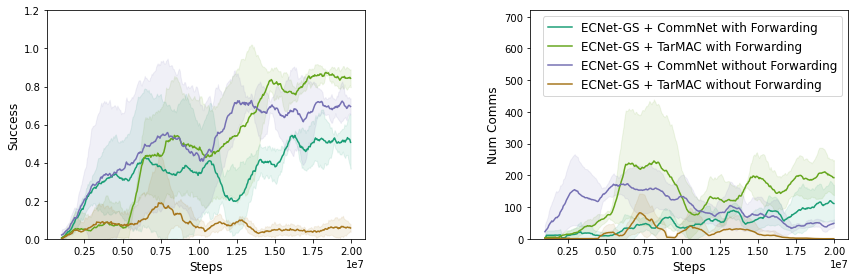}
    \caption{Effect of Message Forwarding in Predator-Prey using ECNet-GS}
    \label{fig:meet_forwarding}
\end{figure*}

In~\Cref{fig:meet_forwarding}, we plot the performance of runs using ECNet-GS and penalty 1e-3, with and without message forwarding. Once again, we see that message forwarding is effective at stabilizing training: most notably, TarMAC with message forwarding is the best-performing run while TarMAC without message forwarding does not train. However, the positive effect is not universal: CommNet performs better without message forwarding.

\subsection{Secret Pairs}

\begin{figure*}[ht]
    \centering
    \includegraphics[width=0.4\linewidth]{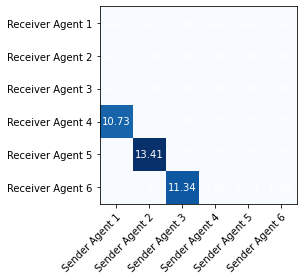}
    \caption{Number of Messages Sent between Agents in Secret-Pairs by a ECNet-GS Agent. We see that the agents only the agents that have the secret (agents 1 - 3) send messages and they only communicate with other agent with the same id (agents 4-6).}
    \label{fig:secret_pairs_heat}
\end{figure*}

We next investigate whether it is possible to learnt pairwise gating by testing our models in the Secret Pairs environment. We found it helpful to make two changes to the models. First, we set the communication gates to be on when not using the communication penalty. Second, in the case of Gumbel-Softmax, we use a higher temperature value of 0.5.

We find that our best-performing model, ECNet-CS with CommNet, is able to achieve reasonable task performance while reducing communication. In order to understand the learnt communication pattern better, we plot the number of messages sent between each pairs of agents in ~\Cref{fig:secret_pairs_heat}. We see that the model achieves the correct pattern of connectivity: only the agents that are given the secret communicate (agents 1-3) and they only send messages to the other agent with the same ID (agents 4-6).

We note that there are limitations to our current pairwise gating network. Specifically, since the gating module only receives the current observation, it cannot use the history of past timesteps to make the current gating decision.

\subsection{Dynamic Cooperative Navigation}

\begin{table*}[ht]
    \small
    \centering
    \begin{tabular}{lrrrrr}
    \toprule
    & \multicolumn{1}{c}{\textbf{Episode Length}} & \multicolumn{1}{c}{\textbf{Message Probability}} \\ \midrule
    {CommNet} & {14.80$\pm$1.42} & {1.00$\pm$0.00} \\[0.05in] \midrule
    {ECNet-Reinforce, Penalty=0.2} & {16.35$\pm$0.43} & {0.29$\pm$0.15} \\[0.05in]
    {ECNet-Reinforce, Penalty=0.4} & {17.85$\pm$0.82} & {0.16$\pm$0.01} \\[0.05in]
    {ECNet-GS, Penalty=0.1} & {18.72$\pm$0.77} & {0.30$\pm$0.03} \\[0.05in] \midrule
    {Random Gate=0.5} & {18.65$\pm$0.3} & {0.50$\pm$0.00} \\[0.05in]
    {Random Gate=0.3} & {22.17$\pm$0.92} & {0.30$\pm$0.00} \\[0.05in]
    {Random Gate=0.15} & {29.40$\pm$1.31} & {0.15$\pm$0.00} \\[0.05in] \midrule
    {ETC-Net} & {16.5$\pm$6.03} & {0.46} \\[0.05in]
    \bottomrule
    \end{tabular}
    \caption{Episode Length and Number of Messages in Dynamic Cooperative Navigation.}
    \label{tab:dynamic_nav_brief}
\end{table*}

Finally, we compare ECNet to ETC-Net on the Dynamic Cooperative Navigation environment. In order to investigate the need for learnt communication decisions in this environment, we also train models that use random gating with various protocols. Our results are shown in \cref{tab:dynamic_nav_brief}. First, we note that our best-performing model, EC-Reinforce with a penalty of 0.2, outperforms ETC-Net on both performance and communication usage. In particular, its communication probability of 0.29 is considerably lower than ETC=Net's probability of 0.46. Second, we observe that the random gating policy with probability of 0.5 has similar communication overhead as ETC-Net and only performs slightly worse. This indicates that it is not actually necessary to train a gating policy to achieve ETC-Net's reduction; instead, you can use random dropout during training time and allow the system to adapt. Third, at lower communication probabilities, EC-Net's training method is clearly necessary: EC-Net performs much better than the corresponding random gating policies at probabilities of 0.3 and 0.15. Thus, while ETC-Net could possibly be replaced with random gating, our method ETC-Net is still significantly superior.

\section{Conclusion}

In this work, we have demonstrated that it is possible to minimize communication while maximizing performance in MARL with a simple method: directly optimizing both the task reward and a communication penalty. Further, we have shown that two techniques, multitask training and message forwarding, can significantly increase performance and stabilize training. Finally, we have shown that we can also learn the pairwise connectivity pattern required to solve a task. Our method opens several exciting possibilities for future research. An interesting future thread is to explore whether it is possible to train a single model to handle multiple communication penalty settings, allowing for deployments with different penalties without retraining.

\textbf{Impact} Multiagent Reinforcement Learning (MARL) may be used in systems that have positive and negative social effects. ECNet could be deployed in any such system. The core benefit of ECNET, reducing communication while retaining performance, will make MARL more accessible and reduce energy consumption. It will enable MARL to be deployed in regions with poor network infrastructure, where sending messages may be expensive. Therefore, it could allow applications such as delivery of supplies or mapping to be expanded.

\bibliographystyle{plainnat}
\bibliography{references}

\appendix
\section{Appendix}

\subsection{Complete Results}

In this section, we report results for all the configurations we tested. In \textbf{Secret} and \textbf{Predator-Prey}, we compare the performance and communication levels of ECNet to the baseline networks and IC3Net. In \textbf{Predator-Prey}, we also illustrate the ability of ECNet to make different trade-offs between performance and communication by testing with low and high penalty settings. Our \textbf{Secret-Pairs} results focus on the ability of ECNet to learn pairwise gating.

\begin{table*}[ht]
    \small
    \centering
    \begin{tabular}{lrrrrr}
    \toprule
    & \multicolumn{1}{c}{\textbf{Penalty}} & \multicolumn{1}{c}{\textbf{Sender Return}} & \multicolumn{1}{c}{\textbf{Receiver Return}} & \multicolumn{1}{c}{\textbf{Sender Comms}} & \multicolumn{1}{c}{\textbf{Receiver Comms}} \\ \midrule
    \textbf{{Baselines}}\\[0.05in]\midrule
    {CommNet} & {0.0} & {19.93$\pm$0.12} & {18.98$\pm$0.12} & {76.00$\pm$0.00} & {76.00$\pm$0.00} \\[0.05in]
    {TarMAC} & {0.0} & {19.23$\pm$0.68} & {18.35$\pm$0.51} & {76.00$\pm$0.00} & {76.00$\pm$0.00} \\[0.05in]
    {TarMAC-Sigmoid} & {0.0} & {19.72$\pm$0.42} & {17.93$\pm$0.71} & {76.00$\pm$0.00} & {76.00$\pm$0.00} \\[0.05in]
    \midrule\textbf{{IC3Net}}\\[0.05in]\midrule
    {CommNet} & {0.0} & {19.10$\pm$0.90} & {18.13$\pm$0.78} & {51.67$\pm$10.30} & {39.78$\pm$5.36} \\[0.05in]
    {TarMAC} & {0.0} & {16.43$\pm$5.04} & {15.69$\pm$4.16} & {62.69$\pm$4.31} & {64.72$\pm$4.91} \\[0.05in]
    {TarMAC-Sigmoid} & {0.0} & {19.24$\pm$0.50} & {18.08$\pm$0.31} & {63.88$\pm$3.93} & {46.11$\pm$8.93} \\[0.05in]
    \midrule\textbf{{ECNet-REINFORCE}}\\[0.05in]\midrule
    {CommNet} & {0.1} & {18.13$\pm$0.62} & {15.50$\pm$1.07} & {3.35$\pm$0.25} & {0.04$\pm$0.06} \\[0.05in]
    {TarMAC} & {0.1} & {17.25$\pm$2.88} & {12.62$\pm$1.40} & {17.42$\pm$8.54} & {0.21$\pm$0.30} \\[0.05in]
    {TarMAC-Sigmoid} & {0.1} & {19.82$\pm$0.27} & {17.12$\pm$1.58} & {6.47$\pm$5.50} & {0.00$\pm$0.00} \\[0.05in]
    \midrule\textbf{{ECNet-GS}}\\[0.05in]\midrule
    {CommNet} & {0.01} & {19.21$\pm$1.33} & {12.32$\pm$2.90} & {2.45$\pm$0.68} & {0.00$\pm$0.00} \\[0.05in]
    {TarMAC} & {0.01} & {18.61$\pm$1.57} & {4.58$\pm$1.90} & {1.79$\pm$1.65} & {0.00$\pm$0.00} \\[0.05in]
    {TarMAC-Sigmoid} & {0.01} & {19.98$\pm$0.03} & {12.68$\pm$2.57} & {3.38$\pm$1.54} & {0.07$\pm$0.10} \\[0.05in]
    \bottomrule
    \end{tabular}
    \caption{Task Return and Number of Messages in \textbf{Secret}. With each network architecture, we compare our method, ECNet, to baselines that do not use gating and IC3Net, which uses gating but does not use a communication penalty. We find that our best-performing method, ECNet-Reinforce with TarMAC-Sigmoid performs on par with prior methods while using fewer than 7 sender messages and no receiver messages. In comparison, both the non-gating baselines and IC3Net send over 50 messages.}
    \label{tab:secret_results}
\end{table*}

\begin{table*}[ht]
    \small
    \centering
    \begin{tabular}{lrrrr}
    \toprule
    & \multicolumn{1}{c}{Penalty} & \multicolumn{1}{c}{Success} & \multicolumn{1}{c}{Return} & \multicolumn{1}{c}{Comms} \\
    \midrule\textbf{{Baselines}}\\[0.05in]\midrule
    {Independent} & {0.0} & {0.14$\pm$0.06} & {19.81$\pm$1.27} & {0.00$\pm$0.00} \\[0.05in]
    {CommNet} & {0.0} & {0.79$\pm$0.07} & {45.31$\pm$3.46} & {711.00$\pm$0.00} \\[0.05in]
    {TarMAC} & {0.0} & {0.81$\pm$0.07} & {46.15$\pm$3.46} & {711.00$\pm$0.00} \\[0.05in]
    {TarMAC-Sigmoid} & {0.0} & {0.66$\pm$0.36} & {41.24$\pm$12.39} & {711.00$\pm$0.00} \\[0.05in]
    \midrule\textbf{{ECNet-REINFORCE Low Penalty}}\\[0.05in]\midrule
    {CommNet} & {0.005} & {0.30$\pm$0.16} & {27.89$\pm$7.79} & {60.37$\pm$50.64} \\[0.05in]
    {TarMAC} & {0.005} & {0.66$\pm$0.16} & {40.86$\pm$6.97} & {71.96$\pm$38.31} \\[0.05in]
    {TarMAC-Sigmoid} & {0.005} & {0.50$\pm$0.26} & {36.96$\pm$12.76} & {31.59$\pm$32.24} \\[0.05in]
    \midrule\textbf{{ECNet-REINFORCE High Penalty}}\\[0.05in]\midrule
    {CommNet} & {0.1} & {0.37$\pm$0.20} & {33.84$\pm$11.28} & {6.17$\pm$4.63} \\[0.05in]
    {TarMAC} & {0.1} & {0.31$\pm$0.19} & {27.38$\pm$8.91} & {4.09$\pm$4.25} \\[0.05in]
    {TarMAC-Sigmoid} & {0.1} & {0.32$\pm$0.34} & {29.87$\pm$14.49} & {3.00$\pm$4.51} \\[0.05in]
    \midrule\textbf{{ECNet-GS Low Penalty}}\\[0.05in]\midrule
    {CommNet} & {0.001} & {0.50$\pm$0.08} & {37.46$\pm$6.19} & {89.14$\pm$50.69} \\[0.05in]
    {TarMAC} & {0.001} & {0.83$\pm$0.05} & {46.82$\pm$2.10} & {182.15$\pm$60.27} \\[0.05in]
    {TarMAC-Sigmoid} & {0.001} & {0.47$\pm$0.27} & {35.32$\pm$12.18} & {187.59$\pm$135.95} \\[0.05in]
    \midrule\textbf{{ECNet-GS High Penalty}}\\[0.05in]\midrule
    {CommNet} & {0.01} & {0.59$\pm$0.08} & {44.03$\pm$2.97} & {25.16$\pm$6.36} \\[0.05in]
    {TarMAC} & {0.01} & {0.21$\pm$0.07} & {22.42$\pm$2.36} & {6.75$\pm$9.79} \\[0.05in]
    {TarMAC-Sigmoid} & {0.01} & {0.38$\pm$0.28} & {34.61$\pm$12.72} & {44.12$\pm$36.63} \\[0.05in]
    \bottomrule
    \end{tabular}
    \caption{Task Performance and Number of Messages in \textbf{Predator-Prey}. We experiment with low and high penalty settings of ECNet in order to evaluate the ability of our models to make different trade-offs between task success and communication.}
    \label{tab:pred_prey_results}
\end{table*}

\begin{table*}[ht]
    \small
    \centering
    \begin{tabular}{lrrrrr}
    \toprule
    & \multicolumn{1}{c}{\textbf{Penalty}} & \multicolumn{1}{c}{\textbf{Sender Return}} & \multicolumn{1}{c}{\textbf{Receiver Return}} & \multicolumn{1}{c}{\textbf{Sender Comms}} & \multicolumn{1}{c}{\textbf{Receiver Comms}} \\
    \midrule\textbf{{ECNet-REINFORCE Pairwise}}\\[0.05in]\midrule
    {CommNet} & {0.1} & {20.00$\pm$0.00} & {15.04$\pm$6.72} & {11.11$\pm$10.16} & {2.20$\pm$5.62} \\[0.05in]
    {TarMAC} & {0.1} & {20.00$\pm$0.00} & {12.42$\pm$7.96} & {2.26$\pm$5.04} & {1.72$\pm$5.30} \\[0.05in]
    {TarMAC-Sigmoid} & {0.1} & {19.89$\pm$0.24} & {13.25$\pm$7.81} & {7.60$\pm$10.41} & {2.90$\pm$6.95} \\[0.05in]
    \midrule\textbf{{ECNet-GS Pairwise}}\\[0.05in]\midrule
    {CommNet} & {0.01} & {20.00$\pm$0.00} & {15.96$\pm$5.67} & {10.73$\pm$9.40} & {6.19$\pm$8.41} \\[0.05in]
    {TarMAC} & {0.01} & {20.00$\pm$0.00} & {11.97$\pm$8.24} & {0.00$\pm$0.00} & {0.00$\pm$0.00} \\[0.05in]
    {TarMAC-Sigmoid} & {0.01} & {20.00$\pm$0.00} & {13.28$\pm$7.84} & {3.80$\pm$8.50} & {1.95$\pm$5.84} \\[0.05in]
    \bottomrule
    \end{tabular}
    \caption{Task Return and Number of Messages in \textbf{Secret Pairs}. We find that ECNet is able to obtain fairly good reward while minimizing communication.}
    \label{tab:secret_pairs_results}
\end{table*}
\FloatBarrier

\subsection{Computational Resources}
Each of our experiments runs on a server with an Nvidia Tesla GPU and 18 Intel Xeon CPU cores. The GPU is used for model inference during rollouts and for training. The CPU cores are used for parallel environment execution. Each experiment used 20 million environment steps and took under 12 hours.

\subsection{Algorithms}
In this section, we provide sketches of the training algorithms used in ECNet-GS and ECNet-REINFORCE.

\begin{algorithm}[H]\linespread{1.35}\selectfont
\While{Total Timesteps < Num Timesteps}{
    Sample trajectories $s^t_i,a^t_i,r^t_i$ using the current policy $\pi_\theta$.\\
    Compute returns $R^t_i$.\\
    Compute policy loss $L_pg(\theta)=\sum_i \sum_t -R^t_i\log{\pi_\theta(a^t_i|s^t_i)}$.\\
    Compute communication penalty loss $L_{comm}(\theta)=\sum_i \sum_t \pi_\theta(c^t_i|s^t_i)$.\\
    Update policy parameters $\theta \longleftarrow \theta - \alpha \delta_{\theta}(L_pg(\theta) + \lambda L_{comm}(\theta))$.
}
\caption{ECNet-GS}
\end{algorithm}

\begin{algorithm}[H]\linespread{1.35}\selectfont
\While{Total Timesteps < Num Timesteps}{
    Sample trajectories $s^t_i,a^t_i,r^t_i$ using the current policy $\pi_\theta$.\\
    Add communication penalty to rewards: $r^t_i\longleftarrow r^t_i - \lambda c^t_i$.\\
    Compute returns $R^t_i$.\\
    Compute shared returns $SR^t=\frac{1}{N}\sum_i R^t_i$.\\
    Compute policy loss $L_pg(\theta)=\sum_i \sum_t -R^t_i\log{\pi_\theta(a^t_i|s^t_i)}$.\\
    Compute communication penalty loss $L_{comm}(\theta)=\sum_i \sum_t -SR^t_i\log{\pi_\theta(c^t_i|s^t_i)}$.\\
    Update policy parameters $\theta \longleftarrow \theta - \alpha \delta_{\theta}(L_pg(\theta) + \lambda L_{comm}(\theta))$.
}
\caption{ECNet-REINFORCE}
\end{algorithm}

\end{document}